\pdfoutput=1

\documentclass[11pt]{article}

\usepackage[final]{acl}

\usepackage{times}
\usepackage{latexsym}
\usepackage{longtable}
\usepackage{makecell} 
\usepackage{svg}

\usepackage{anyfontsize}

\usepackage[T1]{fontenc}

\usepackage[utf8]{inputenc}

\usepackage{microtype}

\usepackage{inconsolata}

\usepackage{graphicx}

\title{A Survey on Multilingual Mental Disorders Detection\\
from Social Media Data}

\author{Ana-Maria Bucur\textsuperscript{1,2,3}, \textbf{Marcos Zampieri}\textsuperscript{4}, Tharindu Ranasinghe\textsuperscript{5}, Fabio Crestani\textsuperscript{1}\\
\textsuperscript{1}Università della Svizzera italiana, Switzerland\\
\textsuperscript{2}Interdisciplinary School of Doctoral Studies, University of Bucharest, Romania\\
\textsuperscript{3}PRHLT Research Center, Universitat Politècnica de València, Spain\\ 
\textsuperscript{4}George Mason University, USA \textsuperscript{5}Lancaster University, UK \\
\normalsize\texttt{ana-maria.bucur-cosma@usi.ch}
}

\begin{document}
\maketitle
\begin{abstract}

The increasing prevalence of mental disorders globally highlights the urgent need for effective digital screening methods that can be used in multilingual contexts. Most existing studies, however, focus on English data, overlooking critical mental health signals that may be present in non-English texts. To address this gap, we present a survey of the detection of mental disorders using social media data beyond the English language. We compile a comprehensive list of 108 datasets spanning 25 languages that can be used for developing NLP models for mental health screening. In addition, we discuss the cultural nuances that influence online language patterns and self-disclosure behaviors, and how these factors can impact the performance of NLP tools. Our survey highlights major challenges, including the scarcity of resources for low- and mid-resource languages and the dominance of depression-focused data over other disorders. By identifying these gaps, we advocate for interdisciplinary collaborations and the development of multilingual benchmarks to enhance mental health screening worldwide.

\end{abstract}

\section{Introduction}

Nearly half of the global population risks developing at least one mental disorder by age 75 \cite{mcgrath2023age}. Many individuals avoid seeking help due to stigma, which varies across cultures and is shaped by cultural norms, religious beliefs, and social attitudes \cite{ahad2023understanding}. Due to several challenges, including the stigma surrounding mental health, limited access in certain areas, economic factors, and a high global demand for services, the World Health Organization advocates for improved delivery of mental health services, including digital technologies to deliver remote care.\footnote{\href{https://www.who.int/news/item/17-06-2022-who-highlights-urgent-need-to-transform-mental-health-and-mental-health-care}{https://www.who.int/news/item/17-06-2022-who-highlights-urgent-need-to-transform-mental-health-and-mental-health-care}}
Furthermore, integrating remote screening tools and culturally adapted digital interventions is crucial \cite{bond2023digital}, as remote screening can detect language patterns linked to mental disorders from short essays \cite{rude2004language}, text messages \cite{nobles2018identification}, or social media \cite{eichstaedt2018facebook}. However, developing effective screening tools for mental disorders heavily relies on the availability of data. Therefore, in this work, we present the datasets that can be used to train models for screening in languages other than English.

The first well-known study on the detection of mental disorders using social media was conducted by \citet{de2013predicting}. Subsequent research has shown that the language used on Facebook can predict future depression diagnoses found in medical records, indicating that social media data could serve as a valuable complement to depression screening \cite{eichstaedt2018facebook}. Methods used for social media screening focus mainly on English data \cite{skaik2020using,harrigian2021state}, and workshops and shared tasks addressing NLP applications to mental health, such as eRisk \cite{parapar2024overview,crestani2022early}, CLPsych \cite{chim2024overview} and LT-EDI \cite{kayalvizhi2023overview} also primarily use English data.

Current NLP models face major limitations in mental disorders detection in languages other than English. Studies show that cultural differences shape how mental disorders are expressed in language \cite{de2017gender,
aguirre2021qualitative,rai2024key}, which means that markers predictive in English often do not generalize well across different cultures \cite{aguirre2021gender,abdelkadir2024diverse}. Even one of the best predictors of depression in language, the use of the first person pronoun ``I'' \cite{rude2004language}, for example, has different degrees of association with the severity of depression across different demographic groups \cite{rai2024key}. This suggests that markers of mental disorders in language are not universal. In addition, self-disclosure rates vary across cultures; collectivist cultures tend to exhibit lower self-disclosure rates than individualist cultures in online settings \cite{tokunaga2009high}. Furthermore, non-native English speakers tend to use their native language for more intimate self-disclosures on social media, with higher rates of negative disclosure compared to posts in English \cite{tang2011tale}. This could have substantial implications for English-based social media screening tools, as they may overlook signals of mental disorders that are present in posts written in languages other than English.

Recently, there have been efforts to develop detection models that focus on languages other than English, such as Portuguese \cite{santos2024setembrobr}, German \cite{zanwar2023smhd}, Arabic \cite{almouzini2019detecting}, and Chinese \cite{zhu2024sentiment}. There have also been shared tasks specifically designed to address these issues, such as MentalRiskES \cite{marmol2025overview}, which focuses on the early detection of depression, suicide, and eating disorders in Spanish. To further contribute to these important efforts, we present the first survey on the detection of mental disorders from social media data beyond English. This survey aims to promote the development of multilingual NLP models that account for cross-cultural and cross-linguistic differences in online language. 

This paper makes the following contributions:

\begin{enumerate}

    \item We provide a comprehensive list of multilingual mental health datasets that capture linguistic diversity and can be used for developing multilingual NLP models.\footnote{We make the list publicly available, and we will continuously update it: \href{https://github.com/bucuram/multilingual-mental-health-datasets-nlp}{https://github.com/bucuram/multilingual-mental-health-datasets-nlp}}

    \item We discuss cross-cultural and cross-language differences in the manifestations of mental disorders in social media.

    \item We identify and describe several research gaps and future directions in the detection of mental disorders using online data beyond English.

\end{enumerate}

\section{Related Surveys}

In this section, we analyze related surveys on the analysis of mental disorders from social media data. \citet{calvo2017natural} is considered one of the first comprehensive surveys, presenting the datasets and NLP techniques used for mental health status detection and intervention. The survey explores research on various mental health conditions and states, including depression, mood disorders, psychological distress, and suicidal ideation, specifically in non-clinical texts such as user-generated content from social media and online forums. Similarly, recent surveys \cite{skaik2020using,harrigian2021state,rissola2021survey,zhang2022natural,garg2023mental,bucur-etal-2025-datasets,bucur2024state} present the datasets, features, and models used to detect mental disorders from online content, with a primary focus on English language data. \citet{dhelim2023detecting} and \citet{bucur2024state} focus on studies that were published during the COVID-19 pandemic that address mental well-being, loneliness, anxiety, stress, PTSD, depression, suicide, and other mental disorders. In addition to these surveys, \citet{chancellor2020methods} provides a critical review of the study design and methods used to predict mental health status, along with recommendations to improve research in this field. 

Our paper fills an important gap in the literature by offering the first comprehensive survey of research on detecting mental disorders in languages other than English. The most related survey to ours is the one conducted by \citet{garg2024towards}, which focuses solely on low-resource languages and only discusses datasets in Thai, Bengali, Hindi, Japanese, and Korean. In contrast, our survey has a broader scope, covering research on a variety of languages, regardless of their resource availability. Our work complements previous surveys by providing an important data-centric foundation for researchers seeking to deploy and validate NLP methods in non-English contexts.

\section{Mental Disorders Detection Tasks Overview}
\label{sec:tasks}

\begin{figure}[t!]
	\centering
		\includesvg[scale=0.25]{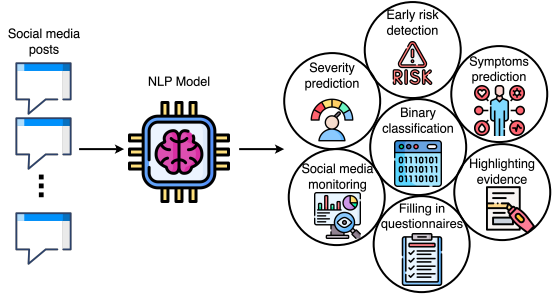}
	\caption{Overview of tasks related to detecting mental health problems from social media.}
	\label{fig:tasks}
\end{figure}

In this section, we discuss common tasks related to predicting mental disorders, emphasizing research available in non-English languages where applicable. The prediction of mental disorders via social media typically involves a supervised \textit{binary classification} task, in which online posts are used to determine the presence or absence of disorders (Figure \ref{fig:tasks}). This can be performed at the post level for predicting suicidal ideation \cite{huang2019suicidal} and depression \cite{uddin2019depression,bucur2023matter}, or at the user level for conditions like depression \cite{hiraga-2017-predicting}, anxiety \cite{zarate2023identifying}, and bipolar disorder \cite{sekulic2018not}. User-level prediction can serve as an \textit{early risk prediction task}, facilitating timely assessments \cite{losada-crestani2016,bucur2021early,marmol2023overview}. \textit{Severity prediction} assesses the intensity of conditions such as depression \cite{10.1145/3485447.3512128,KABIR2023107503,Sampath-dataset} or suicide risk \cite{benjachairat2024classification}. Longitudinal analysis of social media helps detect \textit{moments of change} in mental health \cite{tsakalidis2022identifying}. Other tasks focus on \textit{symptom prediction} \cite{10.1145/3578503.3583621,DBLP:journals/corr/abs-2011-06149} and \textit{highlighting evidence} for mental health problems \cite{chim2024overview,varadarajan2024archetypes}. Mental health indicators from the social media timeline of an individual can be used to \textit{fill in validated questionnaires}, with the goal of estimating symptoms of mental disorders that are usually assessed through survey-based methods such as BDI-II\footnote{\href{https://naviauxlab.ucsd.edu/wp-content/uploads/2020/09/BDI21.pdf}{https://naviauxlab.ucsd.edu/wp-content/uploads/2020/09/BDI21.pdf}} for depression assessment \cite{parapar2021overview} or EDE-Q\footnote{\href{https://www.corc.uk.net/media/1273/ede-q_quesionnaire.pdf}{https://www.corc.uk.net/media/1273/ede-q\_quesionnaire.pdf}} for eating disorders \cite{parapar2024overview}. Lastly, \textit{mental health monitoring} uses aggregated detection results to estimate the prevalence of disorders, as demonstrated during the COVID-19 pandemic \cite{cohrdes2021indications}.

Shared tasks have encouraged interdisciplinary collaborations between psychologists and computer scientists. English-based shared tasks, such as eRisk \cite{parapar2024overview}, CLPsych \cite{coppersmith2015clpsych}, and LT-EDI \cite{kayalvizhi2023overview} provided valuable benchmark datasets that the research community continues to use, even beyond the official competitions. MentalRiskES\footnote{\href{https://sites.google.com/view/mentalriskes2025}{https://sites.google.com/view/mentalriskes2025}} \cite{marmol2025overview} is the only shared task focused on detecting mental disorders in languages other than English. MentalRiskES includes tasks such as the detection of depression, anxiety, eating disorders, and suicidal risk in the Spanish language \cite{marmol2023overview}.

\section{Methodology}

\begin{figure}[t!]
	\centering
		\includegraphics[width=0.45\textwidth]{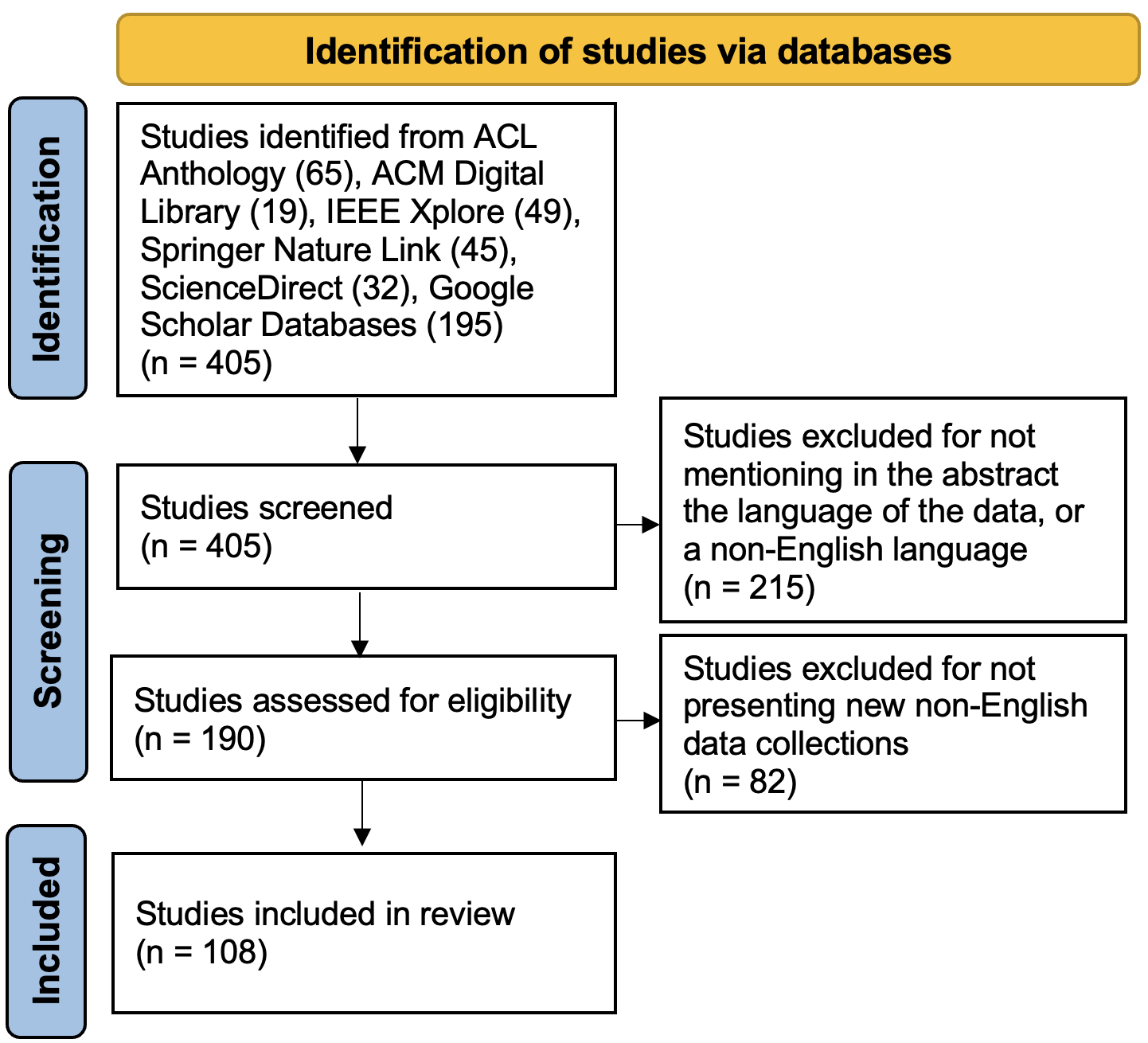}
	\caption{PRISMA flow diagram for our survey.}
	\label{fig:prisma}
\end{figure}

\begin{table*}[t!]
    \fontsize{7.5pt}{7.5pt}\selectfont
    \renewcommand{\arraystretch}{1.3} 
    \begin{tabular}{p{1.4cm}|p{1.5cm}|p{11.5cm}}
        \Xhline{1.5pt}
         \textbf{Language}& \textbf{Resource}  & \textbf{Datasets} \\
         \Xhline{1.5pt}  
         Arabic&  High  &\citet{almouzini2019detecting,alghamdi2020predicting,alabdulkreem2021prediction,musleh2022twitter}, CairoDep \cite{el2021cairodep}, \citet{almars2022attention,maghraby2022modern,baghdadi2022optimized}, Arabic Dep 10,000 \cite{helmy2024depression}, \citet{al2024classification,abdulsalam2024detecting,al2022depression} \\
         \hline 
         Chinese & High & \citet{zhang2014using,huang2015topic,cheng2017assessing,shen2018cross,Wu2018ADA,cao2019latent,wang2019assessing,peng2019multi,huang2019suicidal,li2020automatic}, WU3D \cite{wang2020multimodal}, \citet{yao2020patterns,yang2021fine,chiu2021multimodal,sun2022identification,cai2023depression,li2023mha,guo2023leveraging,wu2023development,lyu2023detecting,yu2023automatic,zhu2024sentiment} \\
          \hline
         French & High  &  \citet{tabak2020temporal} \\
         \hline
         German & High  & \citet{cohrdes2021indications,baskal2022data,tabak2020temporal}, SMHD-GER \cite{zanwar2023smhd}  \\
        \hline
         Japanese & High  & \citet{tsugawa2015recognizing,hiraga-2017-predicting,niimi2021machine,cha2022lexicon,wang2023public}  \\
          \hline
        Spanish& High  & \citet{leis2019detecting}, SAD \cite{lopez2019detecting}, \citet{valeriano2020detection,ramirez2020detection,ramirez2021characterization,villa2023extracting}, MentalRiskES \cite{romero2024mentalriskes}, \citet{cremades2017design,coello2019crosslingual} \\
         \hline
         Brazilian Portuguese& Mid to High & \citet{von2019mining,mann2020see,santos2020searching,de2020machine}, SetembroBR \cite{santos2024setembrobr}, \citet{mendes2024identifying,oliveira2024comparative} \\
         \hline
         Dutch & Mid to High  &\citet{desmet2014recognising,desmet2018online}  \\
         \hline
         Code-Mixed Hindi-English & Mid to High  & \citet{agarwal2021deep} \\
         \hline
         Italian & Mid to High  & \citet{tabak2020temporal} \\
         \hline
         Korean & Mid to High  & \citet{lee2020cross,park2020suicidal,kim2022analysis,kim2022understanding,cha2022lexicon} \\
         \hline
         Polish& Mid to High  & \citet{wolk2021hybrid}  \\
         \hline
         Russian& Mid to High  & \citet{stankevich2019depression,baskal2022data,narynov2020dataset,stankevich2020depression,ignatiev2022predicting}  \\
         \hline
         Turkish & Mid to High  &\citet{baskal2022data}  \\
          \hline
          Bengali& Mid & \citet{uddin2019depression,victor2020machine,kabir2022detection,tasnim2022depressive}, BanglaSPD \cite{islam2022banglasapm}, \citet{ghosh2023attention,hoque2023detecting}, BSMDD \cite{chowdhury2024harnessing} \\
         \hline
         Indonesian & Mid  & \citet{oyong2018natural,yoshua2024depression}  \\
         \hline
         Filipino & Mid & \citet{tumaliuan2024development,astoveza2018suicidal}  \\
         \hline
         Greek & Mid  & \citet{stamou2024establishing}   \\
         \hline
         Hebrew & Mid  & \citet{hacohen2022detection}  \\
         \hline
         Roman Urdu & Mid  & \citet{rehmani2024depression,mohmand2024deep}  \\
         \hline
         Thai& Mid  & \citet{katchapakirin2018facebook,hemtanon2019automatic,kumnunt2020detection,hemtanon2020behavior,wongaptikaseree2020social,hamalainen2021detecting,mahasiriakalayot2022predicting,boonyarat2024leveraging,benjachairat2024classification}   \\
        \hline
         Cantonese & Low  &\citet{gao2019detecting}  \\
         \hline
         Norwegian& Low  & \citet{uddin2022deep,uddin2022depression}  \\
         \hline
         Sinhala& Low$^{*}$ & \citet{rathnayake2021supervised}, EmoMent \cite{atapattu2022emoment}, \citet{herath2024social}  \\
        \hline
    \end{tabular}

    \footnotesize{$^{*}$The classification of Sinhala is based on the work of \citet{ranathunga-de-silva-2022-languages}, who identified the miscategorization of Sinhala in \citet{joshi2020state}'s classification.}
    \caption{Datasets for detecting mental disorders in languages other than English included in the current survey. The availability of resources is based on \citet{joshi2020state}.}
    \label{tab:datasets-small}

\end{table*}

Our primary objective is to catalog the available non-English resources derived from social media for the detection of mental disorders. We conducted a systematic search on major publication databases, including ACL Anthology, ACM Digital Library, IEEE Xplore, Springer Nature Link, ScienceDirect, and Google Scholar. Initially, 405 studies were identified through database searches. After screening the abstracts, 215 papers were excluded because they did not mention the language of the data or mention that the data was in English. Thus, following a review of the main body of the papers, the number of eligible studies was narrowed down to 108, which represents the final count of papers included in this survey. Papers that did not present new data collections in languages other than English were excluded during the screening process. The PRISMA flow diagram is presented in Figure \ref{fig:prisma}. The search terms used in our survey are provided in the Appendix \ref{sec:appendix}.

\section{Multilingual Datasets}
\label{sec:datasets}

In this section, we outline the datasets included in the current survey. 
The languages most frequently represented in these data collections are three high-resource languages: Chinese, Arabic, and Spanish. Although approximately half of the datasets were published in unranked venues, which diminishes their visibility, the other half were published in high-ranking journals and conferences (Figure \ref{fig:datasets} in Appendix \ref{sec:appendix}.

\subsection{Data Sources}

Most of the datasets in English are sourced from Twitter\footnote{All the datasets were collected before Twitter changed its name to X, so we refer to it as ‘Twitter’ in this paper.} and Reddit \cite{harrigian2021state}. Most non-English datasets in this section were also primarily collected from Twitter. However, Reddit was not as widely used for these data collections in non-English contexts. The data presented in this survey come from various populations and regions, and some of the sources are platforms that are exclusive to specific countries, such as Sina Weibo\footnote{\href{https://weibo.com}{https://weibo.com}} used in China, VKontakte\footnote{\href{https://vk.com/}{https://vk.com/}} used in Russia, Pantip\footnote{\href{https://pantip.com/}{https://pantip.com/}} in Thailand, or Everytime\footnote{\href{https://everytime.kr/}{https://everytime.kr/}} in Korea. Complete information about the sources of the data is presented in Table \ref{tab:datasets} in Appendix \ref{sec:appendix}.

The way individuals use social media platforms to share information about themselves varies not only by platform but also by culture. Twitter provides community and safety, helping raise awareness and combat stigma around mental health \cite{berry2017whywetweetmh}. In contrast, Reddit allows for greater anonymity with ``throwaway'' accounts, encouraging users to openly share their experiences in detailed posts on specific subreddits \cite{de2014mental}. This longer format supports post-level mental health analysis \cite{chowdhury2024harnessing}, while Twitter’s shorter posts favor user-level insights, requiring longitudinal data to identify language patterns  \cite{tumaliuan2024development}. Moreover, even when individuals come from the same cultural background and speak the same language, the topics they choose to discuss are heavily influenced by the platform they use. For example, on Sina Weibo, users primarily talk about popular culture, while on Twitter, conversations in Chinese revolve more around politics \cite{zhang2016topical}. These differences, potentially shaped by platform governance and censorship, have direct implications for mental health data collection, as they influence the visibility, framing, and prevalence of mental health-related discourse across different platforms.

\subsection{Languages}

Table \ref{tab:datasets-small} presents all the datasets included in this survey. To classify the availability of resource types, we used the framework proposed by \citet{joshi2020state}, which categorizes resources into six classes (from Rare to High) based on the availability of both labeled and unlabeled NLP data. The languages most frequently represented in the data collections are high-resource languages: Chinese appears in 25 data collections, Arabic is found in 11 datasets, and Spanish is included in 10 datasets. Even if most of the languages covered in the data are from high-, mid to high- and mid-resourced languages, we also have some languages with fewer resources, such as Cantonese, Norwegian and Sinhala. Most of the languages used in the data collections belong to some of the largest language families by number of speakers, specifically the Indo-European, Sino-Tibetan, and Afro-Asiatic language families. When analyzing the language resources based on speaker population and vitality according to Ethnologue \cite{ethnologue26}, we find that all the languages listed in Table \ref{tab:datasets-small} are categorized as large-institutional. This classification indicates that these languages are supported and maintained by institutions beyond their respective communities and have a large number of speakers. The categorization of all listed languages as large-institutional by Ethnologue underscores a robust support system that enhances their viability in NLP applications. However, it also highlights the urgent need to address disparities in resource availability for less-represented languages, promoting inclusivity in NLP research.

\subsection{Mental Disorders}

\begin{figure}[!t]
	\centering
		\includegraphics[width=0.48\textwidth]{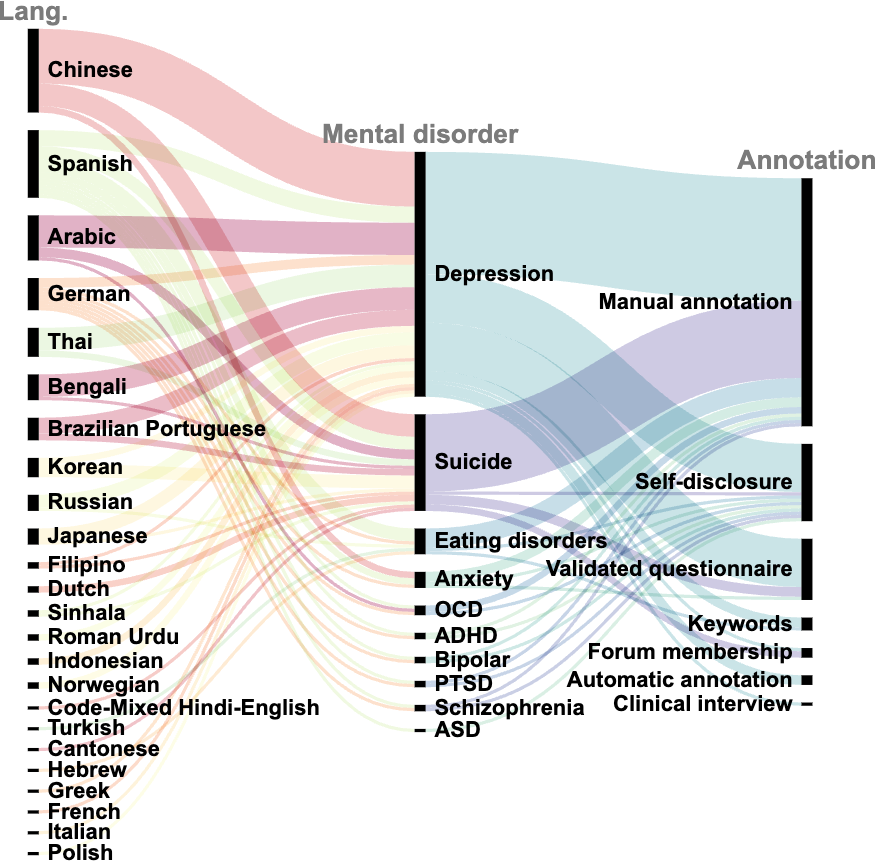}
	\caption{Overview of the mental disorders addressed in each dataset, along with the annotation procedures.}
	\label{fig:annotation}
\end{figure}

Figure \ref{fig:annotation} shows the distribution of mental disorders in different languages within the datasets. Depression is the most common mental disorder and is well-represented in the data. The languages that lack data on depression are Cantonese, Dutch, Hebrew, Hindi, and Turkish. Suicide is another mental disorder that frequently appears in collections. In contrast, the mental health problems that are least represented include eating disorders, obsessive-compulsive disorder (OCD), attention deficit / hyperactivity disorder (ADHD), autism spectrum disorder (ASD), anxiety, bipolar disorder, and schizophrenia. \cite{hazell2022creating}

\subsection{Annotation Procedure}

Most data collections were manually annotated (Figure \ref{fig:annotation}). Manual annotation was carried out by mental health experts or psychologists \cite{narynov2020dataset,de2022can}, graduate students who are native speakers of the language of interest \cite{boonyarat2024leveraging,uddin2019depression}, or nonexpert individuals. However, some datasets do not specify who the annotators were or what guidelines they followed during the annotation process. Most datasets that collect user-level data from online platforms rely on the self-disclosure of mental health statuses. For example, they rely on explicit mentions of diagnoses (e.g. ``I was diagnosed with depression") \cite{tabak2020temporal,villa2023extracting}. The third most common annotation method involves asking social media users to complete validated questionnaires to diagnose mental disorders. The most frequently used survey-based methods include the CES-D \cite{tsugawa2015recognizing,lyu2023detecting}, BDI-II \cite{sun2022identification,stankevich2019depression,ignatiev2022predicting}  or tools specifically designed for certain populations, such as the Thai Mental Health Questionnaire (TMHQ) \cite{katchapakirin2018facebook}. Another reliable annotation approach is conducting clinical interviews to assess mental health problems \cite{wolk2021hybrid}. Less common and noisier annotation methods include identifying posts based on the presence of specific keywords \cite{lopez2019detecting}, by forum membership \cite{agarwal2021deep}, or automatic annotation through another model trained on mental health data \cite{cohrdes2021indications}.

Manual annotation and annotation based on self-disclosure of diagnosis rely on individuals' willingness to disclose personal information on social media. Research shows that collectivist cultures tend to have lower rates of self-disclosure compared to individualist cultures in online contexts \cite{tokunaga2009high}. However, there are notable differences within these cultures. For example, although Koreans are less likely to share personal information, they often provide more in-depth disclosures compared to individuals in the United States \cite{yoo2012exploring}. In addition, self-disclosure annotation relies on individuals explicitly naming their conditions, which may not occur in cultures where mental health is stigmatized. These annotation choices can have significant consequences for model behavior, potentially resulting in systematic false negatives for populations whose expressions of distress do not align with dominant annotation frameworks.

\subsection{Availability of Data Collections}

Out of the 108 datasets listed in Table \ref{tab:datasets-small}, only 23 are publicly available for download without restrictions. This is not surprising, as mental health datasets are often subject to privacy, ethical, and legal concerns, unlike many other NLP domains. These open datasets focus on the detection of depression, suicide, or anorexia in various languages, including Arabic, Bengali, Brazilian Portuguese, Chinese, Hebrew, Hindi, Spanish, Russian, Roman Urdu, and Thai. For 15 of the datasets, access can be obtained by contacting the authors of the respective research papers, while four datasets require users to complete a data agreement to gain access. In addition, four datasets are unavailable due to the sensitive nature of the data. For the remaining datasets, the research papers do not provide any information on data availability. Details about the availability of data collections can be found in Appendix \ref{sec:appendix}, Table \ref{tab:datasets}.

\section{Mental Disorders Detection Approaches}
\label{sec:multilingual}

In this section, we present the methods proposed for the datasets in Section \ref{sec:datasets}. Most approaches are monolingual and specifically target only one non-English language. In Appendix \ref{sec:appendix}, Table \ref{tab:datasets}, we include detailed information on the methods used and evaluation scores for all the datasets.

\paragraph{Classical Monolingual Approaches}~~ Most approaches rely on feature-based text representations such as Bag-of-Words, TF-IDF, and Word2Vec, combined with traditional classifiers like SVM, Logistic Regression \cite{almouzini2019detecting,alghamdi2020predicting,helmy2024depression} or deep learning models \cite{mann2020see,tasnim2022depressive,ghosh2023attention}. These methods are straightforward and can be adapted to any language, making them suitable for the detection task.

\paragraph{Monolingual pre-trained models}~~ With the rise of large pre-trained transformers, language-specific models such as Chinese BERT \cite{yao2024multi}, AraBERT \cite{abdulsalam2024detecting}, German BERT \cite{zanwar2023smhd}, Bangla BERT \cite{chowdhury2024harnessing}, BERTimbau \cite{santos2024setembrobr}, IndicBERT \cite{agarwal2021deep} are being used for detecting mental disorders. However, a significant limitation of these pre-trained transformer-based models is that they are available only for a limited number of high- and mid-resource languages, rendering them impractical for use with low-resource languages.

\paragraph{Multilingual models}~~ Methods developed for multiple languages simultaneously are rare. Some approaches use cross-lingual embeddings and leverage information from languages with more extensive mental health-related resources, such as English, to make predictions on Spanish data \cite{coello2019crosslingual}. \citet{lee2020cross} developed a cross-lingual model for suicidal ideation by translating data from Korean to English and Chinese. Although multilingual pre-trained models like XLM-Roberta and Multilingual BERT are frequently used, they often perform worse than language-specific models, such as BERTimbau and HeBERT \cite{oliveira2024comparative, hacohen2022detection}. This discrepancy suggests that adapting models to specific cultural and linguistic contexts is often more effective than relying solely on general multilingual pre-training. Consequently, relying on a single model for multiple languages can increase the likelihood of errors.

\paragraph{Translation-based}~~ \citet{zahran2025comprehensive} conducted a comprehensive evaluation of LLMs on Arabic data related to depression, suicidal ideation, anxiety, and other mental disorders. The authors showed that LLMs performed better on original Arabic datasets than on data that had been translated into English. Some studies also use data translated from the target language to English \cite{vajrobol2023explainable}. When developing mental health models in languages other than English, certain studies use translations from English to the target language, as seen with Greek \cite{skianis2024leveraging} and various Indian languages \cite{rajderkar2024multilingual}. However, \citet{schoene2025lexicography} has shown that automatically translating suicide-related dictionaries from English to low-resource languages often results in spelling errors and fails to capture the cultural nuances of speakers in the target language.

\paragraph{LLMs}~~ While LLMs have been successfully used in mental health tasks in English \cite{wang2024explainable,shah2025advancing}, research for other languages is limited. For instance, \citet{zahran2025comprehensive} evaluated LLMs on Arabic data and found that slight changes in instructions could harm performance, leading to parsing errors and prompt adherence issues. There are also concerns about increased hallucinations \cite{lauscher2025much} and unsafe responses \cite{wang2024explainable} in languages other than English. Consequently, the use of LLMs in mental health tasks for non-English languages should be approached with caution.

\paragraph{Performance across different languages} As mentioned in Section \ref{sec:tasks}, the detection of mental disorders is primarily approached as a classification task and evaluated using standard metrics such as Accuracy, Precision, Recall, and F1-score. Detailed performance scores can be found in Table \ref{tab:datasets} in Appendix \ref{sec:appendix}. The availability of resources has a significant impact on the performance of supervised or pre-trained language models. High-resource languages, such as Japanese, Chinese, and Arabic, generally yield strong results, indicating that larger and well-annotated datasets contribute to more reliable detection. In contrast, some languages, such as Sinhala, Brazilian Portuguese, and Bengali, show highly variable performance when comparing the detection of depression and suicidal ideation. For other languages included in this survey, such as French, German, Russian, and Italian, the performance in detecting depression is relatively low. Current methods demonstrate unequal effectiveness, which may leave speakers of certain languages without appropriate mental health screening tools. The success of these models is influenced by factors such as language-specific methodologies, the quality of datasets, the availability of resources, and cultural differences in expression. In addition, performance may vary depending on whether the task is binary classification, multi-class classification, or regression. Performance can also vary depending on whether the prediction is based on a single post (post-level classification) or multiple posts from the same user (user-level classification).

\section{Cross-cultural and Cross-language Differences in Mental Health Expression}

Culture influences the sources of distress, how it is expressed, how it is interpreted, the process of seeking help, and the responses of others \cite{kirmayer2001cultural}. In addition, the way people perceive themselves influences their mental health. In Western cultures, there is a strong emphasis on personal narratives, and people tend to express their emotions more openly, a trend that is reflected in online posts \cite{tokunaga2009high}. In contrast, in Asian societies, individuals often internalize their emotional struggles or express them indirectly, influenced by their collectivist values \cite{hu2014relation}. Although negative self-thoughts are a common characteristic of depression, in East Asian contexts, self-criticism is often viewed as a sign of healthy functioning \cite{gotlib2008handbook}.

\paragraph{Symptoms of mental disorders~~}  
Cultural differences in the interpretation of mental health symptoms can lead some individuals to minimize psychological distress, often reporting socially acceptable somatic symptoms such as body pain or fatigue \cite{kirmayer2001cultural}. While somatic symptoms are common across cultures, their expression and understanding can vary. Culturally specific idioms of distress, such as ''nervios''\footnote{Translated as ``nerves'' in English.} in Latin American communities, illustrate this, as they encompass both psychological and somatic symptoms and show a high comorbidity with anxiety and mood disorders \cite{de2000prevalence}. The DSM-V \cite{APA2013} includes cultural concepts of distress to aid clinicians in recognizing diverse expressions of psychological issues. In addition, \citet{rai2025cross} found that Indian social media users on Reddit are more focused on seeking help than on discussing their symptoms or diagnoses.

\paragraph{Mental health expressions in online language~~} Online expression varies between cultures and has been extensively studied among English-speaking individuals from different regions \cite{de2017gender,
aguirre2021qualitative,rai2024key}. When analyzing data from a peer-support mental health community, \citet{loveys-etal-2018-cross} found that manifestations of negative emotions differ between demographic groups. Moreover, \citet{pendse2019cross} found that users in the US, UK, and Canada employed more clinical language to express mental distress compared to users from India, Malaysia, and the Philippines. 

\paragraph{Variation of features across cultures~~} The tendency for self-focused attention, often referred to as ``I''-language, is considered one of the strongest predictors of depression in language \cite{mihalcea2024developments}. As a result, the frequency of the pronoun ``I'' has been used in previous studies as a feature for detecting depression in English. However, it is crucial to carefully consider the applicability of this marker to non-English languages. This association has not been observed in non-Western individuals \cite{rai2024key} or in speakers of Chinese \cite{lyu2023detecting} or Romanian \cite{trifu2024linguistic}. While the pronoun ``I'' serves as an indicator of depression in English, its usage in other languages requires special attention due to linguistic differences. For example, English requires nouns or pronouns to be explicitly included as subjects in sentences. In contrast, some languages, such as Chinese and Romanian, are pro-drop languages, which allow the subject of the action to be omitted \cite{koeneman2019morphology}, resulting in a lower frequency of the personal pronoun ``I'' in these languages.

\paragraph{Mental health metaphors~~}
Indicators of mental disorders are often displayed through metaphors. Depression is often described as weight, pressure, or darkness, and is often portrayed using containment metaphors \cite{charteris2012shattering}. Metaphors are used by individuals to articulate their experiences, and psychologists use them in the therapeutic process \cite{mould2010use}. Mental illness metaphors have been extensively studied in English \cite{charteris2012shattering,lazard2016using} and have been used to predict mental states \cite{shi2021predicting,zhang2021mam}. With the exception of research in Spanish \cite{coll2023metaphor}, there is a lack of resources to understand metaphors of mental illness beyond English.

Considering cultural and multilingual differences is crucial when developing automated methods for predicting mental disorders based on language. These variations may explain the failure of many predictive models to generalize effectively \cite{aguirre2021gender, abdelkadir2024diverse}.

\section{Research Gaps}

In this section, we highlight several research gaps that we hope will be explored in future studies.

\paragraph{Linguistic features used in English do not transfer reliably across languages}
Many commonly used linguistic markers of mental disorders, such as the use of first-person singular pronouns or sentiment polarity, are often considered universal features. However, these markers are actually dependent on language and shaped by cultural contexts, as previous research has indicated \cite{trifan2020understanding,rai2024key}. While many studies have investigated markers of depression in language across cultures, most have focused primarily on individuals from various cultures who communicate in English \cite{aguirre2021gender, abdelkadir2024diverse, rai2025cross}. There is a significant lack of research on how cross-cultural factors influence the expression of mental disorders in individuals’ native languages on social media, as well as how these factors affect the identification of such disorders.

\paragraph{Lack of mental health-related data for low-resource languages} As presented in Section \ref{sec:datasets}, most datasets are often from mid- and high-resourced languages, with the exception of Cantonese, Norwegian, and Sinhala. Currently, many languages remain underrepresented, including high-resource languages like French and mid-to-high-resource languages such as Finnish, Croatian, and Vietnamese. Moreover, there is a lack of datasets in low-resource languages, which may hinder the development of online screening tools for individuals who speak these languages. While some studies have attempted to use automatic translation to build datasets in languages other than English, this approach often fails to accurately capture the cultural nuances of native speakers of the target language \cite{schoene2025lexicography}.

\paragraph{Multilingual approaches} As highlighted in Section \ref{sec:multilingual}, most NLP approaches have focused on processing data in a single target language, with multilingual approaches being almost nonexistent. Most existing NLP models developed for mental disorders detection do not support multiple languages effectively, which limits their applicability in multicultural and multilingual settings where mental health issues may manifest differently. While multilingual and cross-lingual approaches are often proposed as solutions to data scarcity, generalizability failures may stem not only from resource limitations but also from unmodeled cultural variation. 

\paragraph{Annotation transparency and consistency} Although most of the datasets presented in this paper rely on manual annotation for labeling the data related to mental disorders, it is often unclear who performed the annotations. The authors of the research papers should provide specific details about the annotation process, such as whether the annotators are mental health experts or non-experts, if they are native speakers of the target language, and whether they understand the cultural differences in the manifestations of mental disorders. These factors significantly impact the quality and reliability of the data, as understanding cultural nuances is essential in interpreting mental health expressions. For annotations based on questionnaires, it is crucial to use validated questionnaires specific to the target language rather than general English versions. Previous research has demonstrated that incorporating information about cultural idioms of distress in psychological assessments for mental disorders can enhance their validity \cite{cork2019integration}. Therefore, such considerations should also be integrated into the annotation process for mental disorders across cultures.

\paragraph{Explainability} While many mental health studies in English emphasize the importance of explainable approaches \cite{yang2023evaluations,souto2023explainability,yang2023mentalllama}, there is a significant opportunity for applying explainable approaches to non-English languages. This could enhance our understanding of how mental disorders manifest culturally in social media language and provide insights into the diverse expressions of mental disorders.  Currently, few studies have examined model explainability in Bengali \cite{ghosh2023attention} and Thai \cite{vajrobol2023explainable}.

\section{Conclusion}

In this paper, we presented a comprehensive survey of research for mental disorders detection in social media data beyond English. We highlight cross-cultural and multilingual differences in mental health expressions and provide a comprehensive list of datasets that can be used to develop multilingual NLP models for online mental health screening. Our focus was on non-English resources, as most previous research has focused on English \cite{skaik2020using,harrigian2021state}. Lastly, we identified several gaps in current research that we hope will be addressed in future interdisciplinary studies.

\paragraph{Future Directions and Calls to Action} We aim to encourage researchers to develop mental health datasets in languages other than English, fostering interdisciplinary collaborations with experts from psychology and mental health organizations, as seen in successful previous projects like REMO COST Action\footnote{\href{https://projects.tib.eu/remo}{https://projects.tib.eu/remo}}, and PsyMine \cite{ellendorff2016psymine}, which have primarily focused on English. By involving community members, multilingual shared tasks can be organized to identify mental disorders across different languages, inspired by successful SemEval multilingual tasks for offensive language \cite{zampieri-etal-2020-semeval} and emotion detection \cite{muhammad-etal-2025-brighter,muhammad-etal-2025-semeval}. Researchers can work together to annotate data in underrepresented languages while adhering to ethical protocols. By participating in these tasks, members of the ACL community can gain access to datasets that are essential for developing multilingual models. Such initiatives will improve the visibility of multilingual mental disorder detection and encourage further collaborations, providing researchers with more opportunities to address challenges in this field. Researchers can focus on building datasets for underrepresented mental disorders beyond depression, adhering to ethical guidelines and providing transparency in the annotation process \cite{benton2017ethical}. Recent advances in explainability can also be applied to better understand the cultural manifestations of mental disorders.

\section*{Limitations}

Our paper aims to provide a comprehensive survey of cross-cultural language differences and the datasets available for developing multilingual NLP models. We included 108 data collections in this study and carefully reviewed each paper cited in our survey. However, it is possible that we may have overlooked some works that do not explicitly mention in their title or abstract that they focus on languages other than English.

\section*{Ethical Considerations}

\paragraph{Data Collection} We recognize that using online data to identify mental disorders is a promising approach for early screening, but it also presents several ethical challenges \cite{benton2017ethical,chancellor2020methods}. To ensure that research protocols in this area comply with ethical guidelines, researchers must take the following steps: (1) obtain Institutional Review Board (IRB) approval, (2) follow ethical research protocols to protect sensitive data, as outlined by \citet{benton2017ethical}, (3) obtain consent from participants, (4) de-anonymize the data and store it on a secure server. Any further sharing of the data with other researchers must adhere to the same ethical protocols. From our survey of 108 datasets, we found that only 18 received ethical approval from an IRB. In addition, 19 papers indicated that they anonymized the data to protect user privacy. It is concerning that only about 35\% of the papers adhered to ethical practices in their research, highlighting the urgent need for a greater emphasis on ethical standards, especially since ethical disclosures were expected to gradually increase over time \cite{ajmani2023systematic}.

\paragraph{Potential Consequences} Moreover, the ethical implications extend beyond data collection and storage. Researchers should consider the potential consequences of their findings on the populations studied and ensure that their work does not inadvertently stigmatize or harm individuals with mental health disorders. Incorrect predictions can have harmful effects on individuals' lives. For instance, if a system falsely predicts that someone shows signs of mental disorders, it can adversely impact their well-being due to the stigma associated with such labels. This may lead individuals to believe there is something wrong with them, ultimately lowering their self-esteem \cite{chancellor2019taxonomy}. A false negative prediction occurs when the system fails to identify significant signs of distress, preventing the individual from receiving the necessary treatment or interventions. False negative predictions are particularly critical in cases of suicidal ideation, where a person's life may be at risk. \citet{chancellor2019human} critically discuss how subjects are represented in this area of research, highlighting the risk of inadvertently dehumanizing individuals. The language used in mental health-related papers can unintentionally perpetuate stigma, often referring to those involved in data collection as ``sufferers'' of mental disorders while labeling others as ``normal.'' Engaging with the community and stakeholders during the research process can help mitigate these risks and foster a more responsible approach to using online data in mental health research \cite{chancellor2019taxonomy}. 

\paragraph{Model Validity} There are ongoing concerns regarding the construct validity of models trained on data collected from social media, specifically whether these models effectively measure the manifestations of mental disorders \cite{chancellor2020methods}. The datasets used in this survey predominantly rely on manual annotation or labeling through validated questionnaires, which are considered more reliable methods for annotation. However, it is essential to conduct interdisciplinary research and ground the constructs being measured in both theoretical and clinical frameworks. For example, clinical depression (or major depressive disorder) is fundamentally different from merely ``feeling depressed.'' The latter may refer to temporary feelings, while clinical depression encompasses a range of persistent symptoms. These symptoms may include depressed mood, loss of interest in previously enjoyed activities, changes in body weight, sleep disturbances, fatigue, psychomotor agitation or retardation, feelings of guilt, and thoughts of death or suicidal ideation \cite{APA2013}. To be diagnosed with depression, these symptoms must be persistent and significantly impair an individual's ability to function. Prioritizing interdisciplinary collaboration and rigorous validation methods is essential in addressing the complexities of mental health.

\paragraph{Representativeness}  It is important to note that individuals active on social media represent only a subset of the overall population. As a result, there may be differences in how mental disorders are expressed among social media users compared to the general population. Using social media data can introduce bias, as it tends to reflect the experiences of younger and more technologically literate individuals who are more likely to engage with these platforms \cite{chancellor2019taxonomy}. In addition, datasets that include self-disclosure of a mental health diagnosis often come from individuals who are more likely to have sought professional help for their diagnosis and/or treatment. Furthermore, not everyone feels comfortable sharing sensitive information about their mental health online \cite{chancellor2019taxonomy}.

\paragraph{Cultural and Linguistic Variation} Understanding cultural and linguistic variations is crucial when developing automated methods for predicting mental disorders, as they help explain why many predictive models struggle to generalize effectively on data from different demographics \cite{aguirre2021gender,aguirre2021qualitative,abdelkadir2024diverse}. Furthermore, each individual's experience with depression is unique, and it is important to consider their distinct experiences and symptomatology. Algorithmic representations and abstractions play a crucial role in the understanding of mental illness and well-being by providing a framework for generalization \cite{chancellor2019human}. While these simplifications can help identify trends and better understand complex individual experiences, they also risk oversimplifying those experiences. It is important to recognize that generalizing can sometimes lead to misunderstandings regarding the unique nuances of mental health experiences and symptoms. Each person's experience with mental health disorders is unique, and acknowledging this is essential for a deeper understanding of mental health.

\section*{Acknowledgments}

We sincerely appreciate the insightful feedback provided by the anonymous reviewers, which has greatly contributed to the improvement of this paper.

\bibliography{custom}

\appendix

\clearpage

\section{Appendix}
\label{sec:appendix}
\subsection{Methodology details}

The search terms used in our survey are the following: (“mental disorder” OR “mental illness” ” OR “mental health” OR “depression” OR “anxiety” OR “bipolar” OR “ptsd” OR “post traumatic stress disorder” OR “suicide” OR “suicide ideation” OR “ocd” OR “obsessive compulsive disorder” OR “schizophrenia” OR “eating disorder” OR “anorexia” OR “bulimia”) AND “social media” AND (“multilingual” OR the list of languages provided in \citet{joshi2020state}).

\subsection{Rankings of the publication venues for the multilingual datasets}

\begin{figure}[!h]
	\centering
		\includegraphics[width=0.5\textwidth]{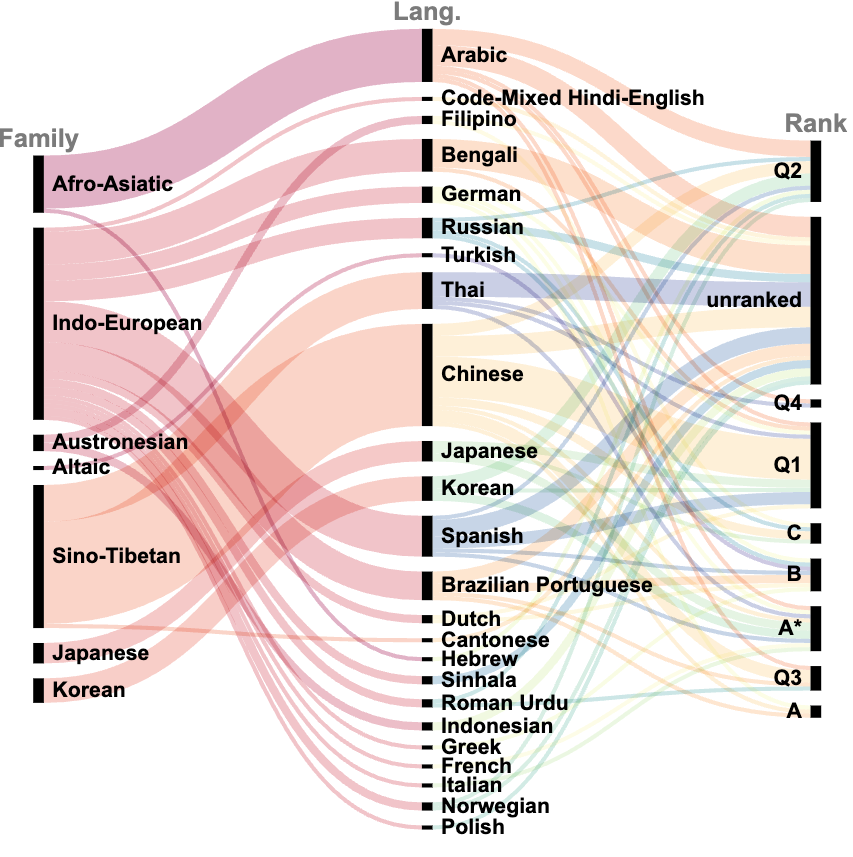}
	\caption{Overview of the languages in the datasets, their language families, and the ranking of their publication venues.}
	\label{fig:datasets}
\end{figure}

Figure \ref{fig:datasets} presents an overview of these languages along with the ranking of the publications in which they appeared. The rankings for conferences are categorized as A$^*$, A, B, and C, following the CORE Rankings Portal.\footnote{\href{https://www.core.edu.au/conference-portal}{https://www.core.edu.au/conference-portal}} For journals, the rankings are classified as Q1, Q2, Q3, and Q4, based on the Journal Citation Reports.\footnote{\href{https://jcr.clarivate.com/}{https://jcr.clarivate.com/}} There are also datasets published in unranked conferences or journals. While about half of the datasets appeared in unranked venues, leading to lower visibility for the research, the other half were published in high-ranking journals and conferences.

\onecolumn

    {\scriptsize 
    \begin{longtable}[c]{p{1.5cm}|p{1.1cm}|p{1.1cm}|p{1.2cm}|p{1.9cm}|p{1.1cm}|p{1.2cm}|p{0.8cm}|p{1cm}|p{1cm}}
    
    \caption{List of Non-English available datasets for mental disorders-related tasks using data posted on online platforms.}\label{tab:datasets}\\

    \textbf{Dataset}& \textbf{Language}& \textbf{Mental disorder} & \textbf{Platform} & \textbf{Annotation Procedure} & \textbf{Label} & \textbf{Dataset Size} & \textbf{Availab.} & \textbf{Method} & \textbf{Performance}\\
    \hline 
    \endfirsthead
    
    \textbf{Dataset}& \textbf{Language}& \textbf{Mental disorder} & \textbf{Platform} & \textbf{Annotation Procedure} & \textbf{Label} & \textbf{Dataset Size} & \textbf{Availab.} & \textbf{Method} & \textbf{Performance}\\
    \hline 
    \endhead
    
    \hline
    \endfoot
    
    \hline
    \endlastfoot

\citet{almouzini2019detecting}&Arabic&depression&Twitter&Self-disclosure &Binary &89 users, 2.7K posts&UNK & Bag-of-Unigrams, Linear SVM & Accuracy: 87.5\%, F1-score: 87.5\%\\
\citet{alghamdi2020predicting} &Arabic&depression&Online forums&Manual annotation &Binary &20K posts &UNK & Lexicon-based & Accuracy: 80.45\%, F1-score: 80.81\%\\
\citet{alabdulkreem2021prediction} &Arabic&depression&Twitter&Manual annotation &Binary &200 users&UNK & Word2Vec, RNN-LSTM & Accuracy: 72\%, F1-score: 69\%\\
\citet{musleh2022twitter} &Arabic&depression&Twitter&CES-D and self-disclosure &Binary, DSM-5 symptoms &4.5K posts &UNK & TF-IDF, RF  & Accuracy: 82.39\%, F1-score: 82.53\%\\
CairoDep \cite{el2021cairodep} &Arabic&depression&Twitter, Reddit, Online forums&Keywords, Manual annotation&Binary &2.4K posts&FREE & AraBERT & Accuracy: 96.93\%, F1-score: 96.92\%\\
\citet{almars2022attention} &Arabic&depression&Twitter&Manual annotation &Binary &6.1K posts&UNK & Attention BiLSTM & Accuracy: 83\%, F1-score: 83\%\\
\citet{maghraby2022modern} &Arabic&depression&Twitter&PHQ-9 &PHQ-9 symptoms &1.2K posts &FREE  &  - & -\%\\
AraDepSu \cite{hassib2022aradepsu} &Arabic&depression, suicidal ideation&Twitter&Manual annotation&Depression, depression with suicidal ideation, or non-depression&20K posts&UNK & MARBERT & Accuracy: 91.20\%, F1-score: 88.75\%\\
Arabic Dep 10,000 \cite{helmy2024depression} &Arabic&depression&Twitter&Manual annotation &Binary &10K posts&FREE  & TF-IDF, RBF SVM & F1-score: 96.6\%\\
\citet{al2024classification} &Arabic&OCD&Twitter&Manual annotation &Binary &8.7K posts&UNK & fastText, RF & F1-score: 80\%\\
\citet{baghdadi2022optimized} &Arabic &suicidal ideation&Twitter&Manual annotation&Binary &2K posts&FREE  & AraBERT &  Accuracy: 96.06\%, F1-score: 95.86\%\\
\citet{abdulsalam2024detecting} &Arabic &suicidal ideation&Twitter&Manual annotation &Binary &5.7K posts&UNK & AraBERT & Accuracy: 91\%, F1-score: 88\%\\
\citet{al2022depression} &Arabic&depression&Twitter&Manual annotation&Binary &6k posts&UNK & TF-IDF, LR & Accuracy: 82\%, F1-score: 81\%\\
\citet{uddin2019depression} &Bengali&depression&Twitter&Manual annotation &Binary &1.1K posts &FREE & GRU & Accuracy: 75.7\%\\
\citet{victor2020machine} &Bengali&depression&Facebook, Twitter&Manual annotation&Binary &30K posts&UNK & TF-IDF, RF & Accuracy: 90\%\\
\citet{kabir2022detection} &Bengali&depression&Facebook&Manual annotation &Depression severity  &5K posts &FREE  & BiGRU & Accuracy: 81\%, F1-score: 81\%\\
\citet{tasnim2022depressive} &Bengali&depression&Facebook&Manual annotation&Binary &7K posts&UNK & BOW, TF-IDF, DT & Accuracy: 97\%, F1-score: 97\%\\
BanglaSPD \citet{islam2022banglasapm}  &Bengali&suicidal ideation&Facebook&Manual annotation&Binary &1.7K posts&UNK & fastText, CNN-BiLSTM & Accuracy: 61\%, F1-score: 61\%\\
\citet{ghosh2023attention} &Bengali&depression&Facebook, Twitter, YouTube&Manual annotation &Binary &15K posts &AUTH  & fastText, BiLSTM-CNN & Accuracy: 94.32\%\\
\citet{hoque2023detecting} &Bengali&depression&Facebook&Manual annotation&Depression severity&2.5K posts&UNK &  XLM-RoBERTa & Accuracy: 61.11\%, F1-score: 60.89\%\\
BSMDD \cite{chowdhury2024harnessing} &Bengali&depression&Reddit, Twitter&Manual annotation &Binary &28K posts&FREE & GPT 3.5 & Accuracy: 97.96\%, F1-score: 98.04\%\\
\citet{von2019mining} &Brazilian Portuguese&depression&Twitter&Self-disclosure &Binary &2.9K users&UNK & Hand-crafted features, SVM & F1-score: 79.8\%\\
\citet{mann2020see}&Brazilian Portuguese&depression&Instagram&BDI &Binary &221 users &UNK & ELMo, ResNet, MLP& F1-score: 79\%\\
\citet{santos2020searching}&Brazilian Portuguese&depression&Twitter&Self-disclosure &Binary &224 users &UNK & TF-IDF, LR & F1-score: 69\%\\
\citet{de2020machine} &Brazilian Portuguese&suicidal ideation&Twitter&Manual annotation &Possibly/ Strongly concerning, Safe to ignore&2.4K posts&UNK & BERT-Portuguese & F1-score: 79\%\\
SetembroBR \cite{santos2024setembrobr}  &Brazilian Portuguese&depression&Twitter&Self-disclosure &Binary &18.8K users &FREE  & BERTimbau & F1-score: 63\%\\
\citet{mendes2024identifying} &Brazilian Portuguese&depression symptoms&Facebook&Manual annotation &Depression symptoms&780 posts&UNK & BERTimbau & Precision: 76.14\%\\
\citet{oliveira2024comparative} &Brazilian Portuguese&suicidal ideation&Twitter&Manual annotation &Binary&3.7K posts&FREE & GPT-4 & F1-score: 98\%\\
\citet{gao2019detecting}&Cantonese&suicidal ideation&Youtube&Manual annotation&Binary &5K posts&UNK & Word2vec, LSTM & Geometric mean of accuracies: 84.5\%\\
\citet{zhang2014using} &Chinese&suicidal ideation&Sina Weibo&SPS&SPS score&697 users&UNK & LIWC, LR &  RMSE: 11\\
\citet{huang2015topic} &Chinese&suicidal ideation&Sina Weibo&Manual annotation &Binary&7.3K posts&UNK & Topic modeling, LibSVM & F1-score: 80.0\%\\
\citet{cheng2017assessing}&Chinese&suicidal ideation&Sina Weibo&Suicide Probability Scale (SPS), DASS-21&Binary&974 users&UNK & LIWC, SVM & AUC: 61\%\\
\citet{shen2018cross} &Chinese&depression&Sina Weibo&Self-disclosure &Binary &1.1K users &UNK & Hand-crafted features, DNN & F1-score: 78.5\% \\
\citet{Wu2018ADA}&Chinese&depression&Facebook&CES-D &Binary &1.4K users &UNK & Word2vec, Hand-crafted features ,RNN & F1-score: 76.9\%\\
\citet{cao2019latent}&Chinese&suicidal ideation&Sina Weibo&Manual checking of self-report and/or appartenence to a suicide-related community&Binary &7K users&DUA & fastText, RNN & F1-score: 90.92\%\\
\citet{wang2019assessing} &Chinese&depression&Sina Weibo&Manual annotation &Depression severity  &13.9K users &UNK & BERT & F1-score: 53.8\%\\
\citet{peng2019multi} &Chinese&depression&Sina Weibo&Manual annotation &Binary &387 users &UNK & TF-IDF, SVM & F1-score: 76.12\%\\
\citet{huang2019suicidal}&Chinese&suicidal ideation&Sina Weibo&Manual annotation &Binary &18.5K posts&UNK & LIWC, Dictionary, LR, DT, SVM & F1-score: 88\%\\
\citet{li2020automatic} &Chinese&depression&Sina Weibo&Self-disclosure &Binary &1.8K users &FREE  & Lexicon-based, RF  & F1-score: 76\%\\
WU3D \cite{wang2020multimodal} &Chinese&depression&Sina Weibo&Depression-related keywords &Binary &32K users &FREE XLNet mebeddings, BiGRU &  & F1-score: 96.85\%\\
\citet{yao2020patterns} &Chinese&depression&Sina Weibo&Manual, automatic annotation &Binary &2.7K users &UNK & - & - \\
\citet{yang2021fine} &Chinese&depression&Sina Weibo&Manual annotation &Depression severity&6.1K posts &AUTH  & BERT-based & F1-score: 65.7\%\\
\citet{chiu2021multimodal} &Chinese, English&depression&Instagram&Depression-related keywords &Binary &520 users &UNK & Multimodal features, Adaboost & F1-score: 83.5\%\\
\citet{sun2022identification} &Chinese&suicidal ideation, depression&Sina Weibo&BDI, SDS, Manual annotation&Binary / Possibly/Strongly concerning, Safe to ignore&203 users, 1.2K posts&UNK & Gradient Boosting & F1-score: 78.90\%\\
\citet{cai2023depression} &Chinese&depression&Sina Weibo&Self-disclosure and manual annotation &Binary &23K users &FREE  & DNN & F1-score: 92.02\%\\
\citet{li2023mha} &Chinese&depression&Sina Weibo&Self-disclosure, manual annotation &Binary &4.8K users &UNK & Multimodal features, DNN & F1-score: 92.78\%\\
\citet{guo2023leveraging} &Chinese&depression&Sina Weibo&Manual annotation &Binary &3.1K users &UNK & Lexicon-based, XGBoost & F1-score: 93.22\%\\
\citet{wu2023development} &Chinese&suicidal ideation&Dcard and PTT&Manual annotation &Risk levels&2K posts&UNK & - & -\\
\citet{lyu2023detecting}&Chinese&depression&Sina Weibo&CES-D&Binary&789 users&AUTH & LIWC, LR & Pearson correlation: 0.33\\
\citet{yu2023automatic} &Chinese&anxiety&Sina Weibo&Self-Rating Anxiety Scale&SAS score&1K users&N/A & LIWC, XGBoost & Pearson correlation: 0.32\\
\citet{zhu2024sentiment} &Chinese&anxiety&Sina Weibo&Manual annotation &Binary &6K posts&UNK & LIWC, Word embeddings CNN & F1-score: 86.13\%\\
\citet{wang2024automatic} &Chinese&depression&Sina Weibo&Manual annotation &Binary &14.8K users&AUTH & Multimodal features, DNN & F1-score: 89.15\%\\
\citet{yao2024multi} &Chinese&depression&Sina Weibo&Manual annotation &Binary &200 users&AUTH & BERT, DNN & Accuracy: 90\% \\
\citet{zhang2024natural}&Chinese&depression&Sina Weibo&Manual annotation &Binary &1.6K users&UNK & Tencent Embeddings, HTN & F1-score: 95.43\%\\
\citet{desmet2014recognising}&Dutch&suicidal ideation&Online forums&Manual annotation &Fine-grained labels&1.3K posts&UNK & BOW, SVM & F1-score: 85.6\%\\
\citet{desmet2018online}&Dutch&suicidal ideation&Online forums&Manual annotation &Fine-grained labels&10K posts&UNK & BOW, Topic modeling, LibSVM & F1-score: 92.69\%\\
\citet{abdelkadir2024diverse}&English, but from different populations&depression&Twitter&Self-disclosure, Manual annotation &Binary &531 users&UNK & Mental
Longformer & F1-score: 62\%\\
\citet{tumaliuan2024development}&Filipino, English&depression&Twitter&PHQ-9&Binary&72 users&AUTH & - & -\\
\citet{astoveza2018suicidal}&Filipino, Taglish&suicidal ideation&Twitter&Manual annotation &Binary &2.1K posts&UNK & BOW, MLP & Accuracy: 77.9\%\\
\citet{cohrdes2021indications}&German&depression&Twitter&Automatic annotation for PHQ-8 symptoms&Binary &88K posts &AUTH  & -- & -- \\
SMHD-GER \cite{zanwar2023smhd} &German&depression, ADHD, anxiety, bipolar, OCD, PTSD, schizophrenia&Reddit&Manual annotation &Labels for multiple disorders &28K posts &DUA  & LIWC, BiLSTM & F1-score: 50.89\%\\
\citet{baskal2022data} &German, Russian, Turkish, English&eating disorders&Reddit, Tumblr&Manual annotation&Binary &3K posts&AUTH & -- & --\\
\citet{tabak2020temporal}&German, French, Italian, Spanish, English&depression&Twitter&Self-disclosure &Binary &5K users &UNK & BOW, BiLSTM & F1-score: 69\%\\
\citet{hacohen2022detection}&Hebrew&anorexia&Online forums&Manual annotation &Binary &200 posts&FREE & Hand-crafted features, RF & Accuracy: 90.63\%\\
\citet{agarwal2021deep} &Code-Mixed Hindi-English&suicidal ideation&Reddit&Subreddit membership&Binary &6.4K posts&FREE  & Indic BERT & Accuracy: 98.54\%\\
\citet{oyong2018natural} &Indonesian&depression&Twitter&Manual annotation&Binary &55 users&UNK & Hand-crafted depression score & F1-score: 50\%\\
\citet{yoshua2024depression}&Indonesian&depression&Twitter&DASS-42&Binary &184 users&UNK & Word2Vec, DT& F1-score: 94\% \\
\citet{tsugawa2015recognizing}&Japanese&depression&Twitter&CES-D, BDI&Binary&209 users&UNK & Hand-crafted features, Topic modeling, SVM & Accuracy: 66\%\\
\citet{hiraga-2017-predicting}&Japanese&depression&Online blogs&Self-disclosure &Binary &101 users &UNK & Part-of-speech, NB & Accuracy: 95.5\% \\
\citet{niimi2021machine}&Japanese&depression&TOBYO&Blog theme &Binary &901 users &UNK & TF-IDF, SVM & F1-score: 96.2\%\\
\citet{wang2023public}&Japanese&suicidal ideation&Twitter&Manual annotation&Binary &30K posts&N/A & --& --\\
\citet{lee2020cross} &Korean&suicidal ideation&Naver Cafe&Membership in a forum&Binary &31K posts&UNK & Word2Vec, RNN & F1-score: 87.49\%\\
\citet{park2020suicidal} &Korean&suicidal ideation&Online forums&Manual annotation &Risk levels&2.7K posts&AUTH & KoBERT & F1-score: 88\%\\
\citet{kim2022understanding} &Korean&suicidal ideation&Twitter&Manual annotation&Binary &20K posts, 414 users&UNK & --& --\\
\citet{kim2022analysis}&Korean&depression&Online forums&PHQ-9, Manual annotation&PHQ-9 score, PHQ-9 symptoms&60 users, 28K posts&UNK & BERT-based & F1-score: 93\%\\
\citet{jung2023suicidality}&Korean&suicidal ideation&Twitter&Manual annotation &Binary &20k posts&UNK & Metadata, word count, XGBoost & F1-score: 83.57\%\\
\citet{cha2022lexicon}&Korean, Japanese, English&depression&Twitter, Everytime&Lexicon-based automatic annotation &Binary &26M posts, 22K posts &AUTH  & BERT-based & F1-score: 99\%\\
\citet{stamou2024establishing} &Modern Greek&depression&Twitter&Self-disclosure &Binary &78 users&AUTH & -- & --\\
\citet{uddin2022depression}&Norwegian&depression&Online forums&Manual annotation &Binary &21.8K posts&UNK & TF-IDF, LSTM& F1-score: 97\%\\
\citet{uddin2022deep} &Norwegian&depression&Online forums&Manual annotation&Binary &30K posts&UNK & Hand-crafted depression features; LSTM & F1-score: 98\%\\
\citet{wolk2021hybrid}&Polish&depression&Facebook, Reddit &Self-disclosure, clinical interview &Binary &262 users &UNK & Hybrid Model; BERT & Accuracy 71\%\\
\citet{rehmani2024depression}&Roman Urdu&depression&Facebook&Manual annotation &Depression severity & 3K posts & AUTH & SVM & F1-score: 70.3\%\\
\citet{mohmand2024deep}&Roman Urdu&depression&Twitter&Keywords-based annotations + Expert review&Depression severity&25K posts&FREE & Transfer learning; BERT & F1-score: 99\% \\
\citet{stankevich2019depression}&Russian&depression&VKontakte&BDI &BDI score&531 users&UNK & Psycholinguistic Markers & F1 Score: 65\%\\
\citet{narynov2020dataset}&Russian&depression&VKontakte&Manual annotation &Binary &34K posts&FREE & -- & -- \\
\citet{stankevich2020depression}&Russian&depression&VKontakte&BDI &BDI score &1.3K users &UNK & -- & --\\
\citet{ignatiev2022predicting}&Russian&depression&VKontakte&BDI&Binary &619 users&DUA & CatBoost & F1 Score: 69\% \\
\citet{rathnayake2021supervised}&Sinhala&depression&Twitter, Facebook&Manual annotation &Binary &1K posts&UNK & KNN & F1-score: 68.5\%\\
EmoMent \cite{atapattu2022emoment} &Sinhala, English&mental illness&Facebook&Manual annotation &mental illness, sadness, suicidal, anxiety/stress, psychosomatic, other, irrelevant&2.8K posts&AUTH & RoBERTa & F1 Score: 76\%\\
\citet{herath2024social}&Sinhala&suicidal ideation&Facebook&Manual annotation &Binary &300 posts&UNK & SVM & F1-score: 76\%\\
\citet{leis2019detecting}&Spanish&depression&Twitter&Self-disclosure, manual annotation &Binary &540 users, 1K posts &FREE  & -- & -- \\
SAD \citet{lopez2019detecting}&Spanish&anorexia&Twitter&Hashtags&Binary &5.7K posts&FREE  & SVM & F1-score: 91.6\%\\
\citet{valeriano2020detection}&Spanish&suicidal ideation&Twitter&Manual annotation &Binary &2K posts&FREE & Word2Vec; LR & F1-score: 79\% \\
\citet{ramirez2020detection}&Spanish&suicidal ideation&Twitter&Manual annotation&Binary &252 users&N/A & Psychological features, LR & F1-score: 80\% \\
\citet{ramirez2021characterization}&Spanish&anorexia&Twitter&Manual annotation &Anorexia, control, under treatment, recovered, doubtful&645 users&N/A & CNN, LR & F1-score: 98\% \\
\citet{villa2023extracting} &Spanish, English&depression, ADHD, anxiety, ASD, bipolar, eating disorders, OCD, PTSD, schizophrenia &Twitter&Self-disclosure &Labels for multiple disorders & 6K users &DUA  & N-Grams; XGBoost & F1-score: 82.4\%\\
MentalRiskES \citet{romero2024mentalriskes}&Spanish&depression, anxiety, suicidal ideation, eating disorders&Telegram&Manual annotation &Binary + suffer + in favour (sf), suffer + against (sa), suffer + other (so) for Depression&1.2K users&AUTH & Social media text; mDeBERTa & F1 Score: 46\%\\
\citet{cremades2017design}&Spanish, English&suicidal ideation&Facebook, Twitter, Blogspot, Reddit, Pinterest&Manual annotation &Binary &97 posts&FREE  & -- & --\\
\citet{coello2019crosslingual}&Spanish, English&depression&Twitter&Self-disclosure &Binary &316 users &FREE  & BOW, SVM & F1 Score: 78\%\\
\citet{katchapakirin2018facebook}&Thai&depression&Facebook&TMHQ&Binary &35 users&UNK & RF & F1 Score: 88.9\%\\
\citet{hemtanon2019automatic}&Thai&depression&Facebook&Manual annotation &Binary &1.5K posts &UNK & SVM & F1 Score: 94\%\\
\citet{kumnunt2020detection}&Thai&depression&Pantip&Hashtags&Binary &31K posts&UNK & CNN-LSTM & F1 Score: 83.1\%\\
\citet{hemtanon2020behavior}&Thai&depression&Facebook&PHQ-9&Binary &160 users&UNK & Social media features & F1 Score: 91.4\%\\
\citet{wongaptikaseree2020social}&Thai&depression&Facebook&TMHQ&Binary &600 users&UNK & -- & -- \\
\citet{hamalainen2021detecting}&Thai&depression&Online blogs&Manual annotation &Binary &900 posts &FREE  & BERT & Accuracy: 77.53\%  \\
\citet{mahasiriakalayot2022predicting} &Thai&depression&Twitter&Manual annotation &Depression symptoms&3.1K posts & UNK & LSTM & F1-score: 81.06\% \\
\citet{boonyarat2024leveraging}&Thai&suicidal ideation&Twitter&Manual annotation &Binary&2.4K posts&FREE & Linguistic features; BERT & F1-score: 93\%\\
\citet{benjachairat2024classification} &Thai&suicidal ideation&Twitter&Manual annotation &C-SSRS Labels&5.1K posts&AUTH & Text features; LSTM & F1-score: 93.88\% \\
\hline

\end{longtable}
}

\clearpage
\twocolumn

\end{document}